\documentclass[a4paper,twoside]{article}

\usepackage{epsfig}
\usepackage{subcaption}
\usepackage{calc}
\usepackage{amssymb}
\usepackage{amstext}
\usepackage{amsmath}
\usepackage{amsthm}
\usepackage{multicol}
\usepackage{pslatex}
\usepackage{apalike}
\usepackage{algorithm2e}
\usepackage[bottom]{footmisc}
\usepackage{times}
\usepackage{latexsym}
\usepackage{graphicx}
\usepackage{amsmath,amssymb,amsfonts}
\usepackage{mathtools}
\usepackage{arydshln}
\usepackage{SCITEPRESS}     % Please add other packages that you may need BEFORE the SCITEPRESS.sty package.

\begin{document}

\title{Mitigating Outlier Activations in Low-Precision Fine-Tuning of Language Models}

% \author{\authorname{First Author Name\sup{1}\orcidAuthor{0000-0000-0000-0000}, Second Author Name\sup{1}\orcidAuthor{0000-0000-0000-0000} and Third Author Name\sup{2}\orcidAuthor{0000-0000-0000-0000}}
% \affiliation{\sup{1}Institute of Problem Solving, XYZ University, My Street, MyTown, MyCountry}
% \affiliation{\sup{2}Department of Computing, Main University, MySecondTown, MyCountry}
% \email{\{f\_author, s\_author\}@ips.xyz.edu, t\_author@dc.mu.edu}
% }

\author{\authorname{Alireza Ghaffari\sup{1}, Justin Yu\sup{1}, Mahsa Ghazvini Nejad\sup{1}, Masoud Asgharian\sup{2}, Boxing Chen\sup{1} and Vahid Partovi Nia\sup{1}}
\affiliation{\sup{1}Huawei Noah's Ark Lab, Montreal, Canada}
\affiliation{\sup{2}Department of Mathematics and Statistics, McGill University, Montreal Canada.}
\email{alireza.ghaffari@huawei.com}
}


\abstract{Low-precision fine-tuning of language models has gained prominence as a cost-effective and energy-efficient approach to deploying large-scale models in various applications. However, this approach is susceptible to the existence of outlier values in activation. The outlier values in the activation can negatively affect the performance of fine-tuning language models in the low-precision regime since they affect the scaling factor and thus make representing smaller values harder.
This paper investigates techniques for mitigating outlier activation in low-precision \textit{integer} fine-tuning of the language models. 
Our proposed novel approach enables us to represent the outlier activation values in 8-bit integers instead of floating-point (\texttt{FP16}) values. The benefit of using integers for outlier values is that it enables us to use operator tiling to avoid performing 16-bit integer matrix multiplication to address this problem effectively. We provide theoretical analysis and supporting experiments to demonstrate the effectiveness of our approach in improving the robustness and performance of low-precision fine-tuned language models.}

\onecolumn \maketitle \normalsize \setcounter{footnote}{0} \vfill

\section{\uppercase{Introduction}}

Language models have achieved remarkable success in various NLP tasks, owing to their ability to capture the intricacies of text data. Fine-tuning large language models, however, often requires substantial computational resources and memory bandwidth that hinder its accessibility to users with limited computational resources. To make large language models accessible and efficient for real-world applications, researchers have explored various techniques for making fine-tuning pre-trained models more efficient on devices with lower computational power. To mitigate these challenges, low-precision fine-tuning has emerged as a promising approach.

Low-precision fine-tuning involves representing the model's weights, activations, and also gradients to lower bit-width representations, such as 8-bit integer or floating-point numbers. This approach reduces memory and computational requirements, making it feasible to fine-tune and deploy large-scale models on resource-constrained devices. However, both in fine-tuning and inference, this approach introduces the problem of outlier activation, where a small number of activations exhibit extreme values, causing numerical instability and degradation in model performance.

\begin{figure}[t]
\centering
\includegraphics[width=0.8\columnwidth]{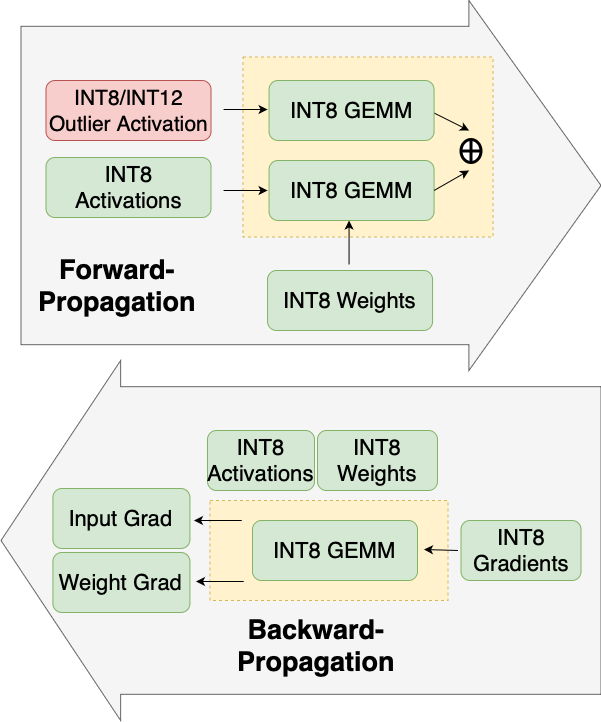}
\caption{Computation flow of proposed linear layers for forward and backward propagation. Integer computation significantly reduces the computational cost of compute-intensive linear layers.}
\label{fig:computation}
\end{figure}

In this paper, we delve into the problem of outlier activation in low-precision fine-tuning of language models. In our proposed approach,  weights, activations, and gradients of all compute-intensive linear layers are represented using integer number formats. 
Instead of quantization approaches used in the literature, we propose a comprehensive approach that uses various hardware design techniques to address this issue effectively. Our contributions are as follows.
\begin{itemize}
    \item We analyze the causes and consequences of outlier activation in low-precision fine-tuning. We find that outlier activations are more important in the forward pass. Thus, we keep all the gradients in the back-propagation of linear layers in 8-bit integers.
    \item Instead of quantizing  the floating point values, we switch the number format of weights, activations, and gradients to an adaptive-integer number format which considers different integer lengths (e.g. \texttt{INT12} or \texttt{INT16}) for activation outliers (less than 5\% of all parameters) and keeps all the other parameters in \texttt{INT8} format. 
    \item Using the advantage of integer number formats, we present a tiling strategy that enables the possibility of using \texttt{int8} GEMM for all the computation of linear layers. Note that such tiling strategy is not easily possible for floating-point number formats such as (\texttt{FP16} and \texttt{FP8})
    \item We provide theoretical analysis on how treating outliers separately helps to preserve the information in low-precision regime.
\end{itemize}

% We analyze the consequences of outlier activation in low-precision fine-tuning.
% We introduce a novel technique for mitigating outlier activation, leveraging gradient clipping to control extreme gradients during training.
% We explore data preprocessing methods to normalize input data and reduce the likelihood of outlier activations.
% We present a targeted weight initialization strategy to align weights with the reduced numerical range in low-precision settings.

In Figure \ref{fig:computation}, we present an overview of our novel linear layers, highlighting the innovative handling of outlier activations in an integer number format while maintaining gradient computation in low-precision integer format. Notably, to the best of our knowledge, this is the first fully \texttt{INT8} linear layer designed to manage outlier features in integer format, while simultaneously preserving gradient calculations in low-precision integer format.

\section{\uppercase{Related Works} }
% With the public reveal of chatbots as an application of Large Language Models (LLM) there is more interest than ever in the development of LLMs. 
The emergence of Large Language Models (LLMs) has revolutionized natural language processing, yet their formidable size poses significant computational challenges for training, fine-tuning, and deployment. To address these challenges, intensive research has focused on quantization techniques, low-precision arithmetic, and compression methods. This Section investigates these approaches and their efficacy in mitigating outlier activation during inference and back-propagation, offering insights into the evolving landscape of techniques designed to make LLMs more efficient and accessible in resource-constrained environments.

\subsection{Handling Outliers in Low-precision Inference}
Most of the research efforts in the literature are focused on studying the effect of outlier activations in the forward propagation i.e. inference. 
% The literature on quantizing the inference normally falls into two categories of Post Training Quantization (PTQ) and Quantized Aware Training (QAT). 
For instance, LLM.int8() proposed by \cite{dettmers2022llm} decompose the outlier activations and their corresponding weights to a separate matrix multiplication that is performed in \texttt{FP16} format while keeping the values that are not outlier in \texttt{INT8} format. They also show that using a threshold is enough for detecting the outlier features.
GPTQ presented by \cite{frantar2022gptq} is a one-shot post-training quantization (PTQ) scheme that is based on approximate second-order information. GPTQ quantizes the weights while keeping the activations in floating point format.
AWQ proposed by \cite{lin2023awq} is another PTQ scheme that focuses on protecting salient activations by applying normalizing scales for weights and activation tensors. These scales are determined by only analyzing activation tensors.
\cite{dettmers2023case} proposed an outlier-dependent quantization scheme called proxy quantization which quantizes the weights corresponding to the outliers into a higher precision number format. Proxy quantization exploits the standard deviation of each layer’s hidden unit weights as a proxy for which dimensions have outlier features.
Outlier channel splitting (OCS) proposed by \cite{zhao2019improving} tackle the problem of outlier features by duplicating channels containing outliers, then halves the channel values.
Morover, norm tweaking is proposed by \cite{li2023norm} to reverse the magnification of outliers by normalization layers i.e. LayerNorm as discovered by \cite{wei2022outlier}.
SmoothQuant proposed by \cite{xiao2023smoothquant} offers an 8-bit quantization scheme for weights and activation. SmoothQuant deals with outlier features by migrating the quantization difficulty from activation to weights using scaling factors for weights and activations.
Furthermore, \cite{dettmers2023spqr} proposed SpQR that isolates outlier weights, which may cause large quantization errors, and stores them in higher precision, an then compresses all other weights to integer format. 
\cite{yuan2023rptq} suggested RPTQ method which reorders the channels with outliers and group them in order to reduce the quantization error.

\begin{figure*}[!t]
\centering
% \vspace{-2cm}
\includegraphics[width=0.4\textwidth,angle=-90]{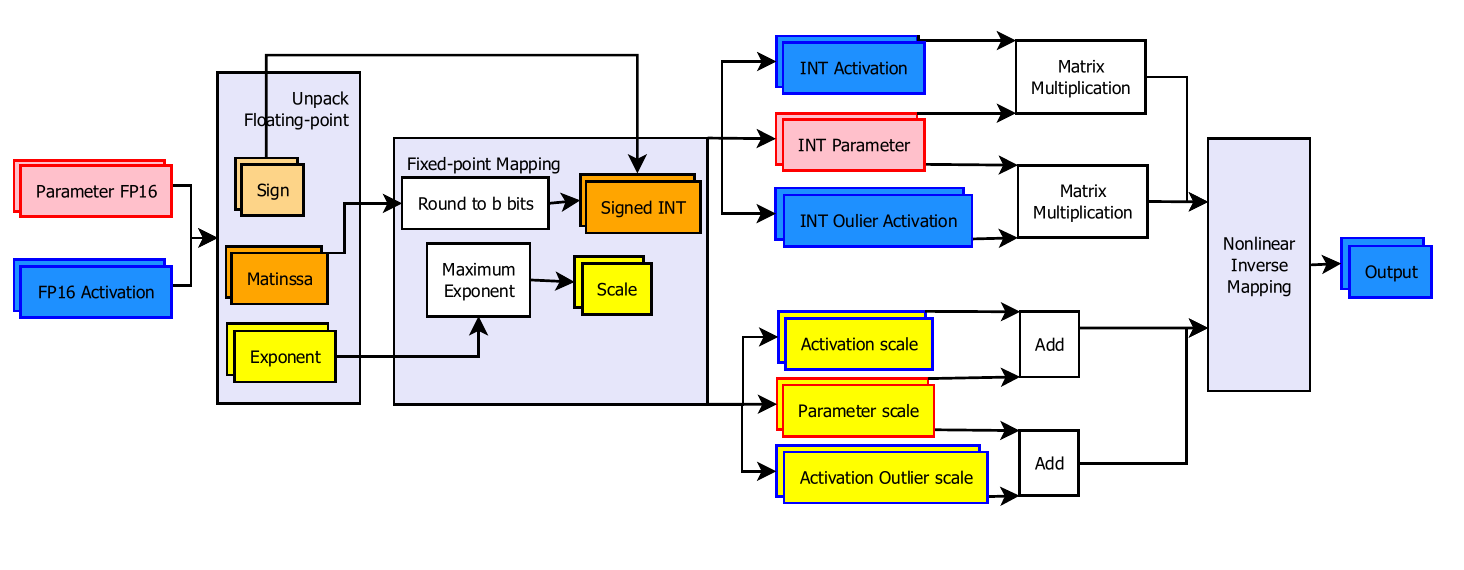}
% \vspace{-5.7cm}
\caption{Inference operations in an integer-only linear layer. The bottom panel shows the linear fixed-point mapping for the input tensors, which can adapt different bit-widths for activation outliers and other parameters in the layer.}
\label{fig:quant}
\end{figure*}

\subsection{Handling Outliers in Low-precision Back-propagation}

Over the past few years, low-precision training of deep learning models has gained popularity in reducing the training cost.
For instance, \texttt{FP16} mixed precision training \cite{micikevicius2017mixed} is nowadays a commonplace methodology to fine-tune language models. Integer data type has also been extensively studied for back-propagation computations by \cite{zhang2020fixed,zhao2021distribution,zhu2020towards,ghaffari2022integer}. Moreover, using higher-bit integer formats such as \texttt{INT12} for both back-propagation and forward propagation is proposed by \cite{hosseini2023towards}. 

Nonetheless, the majority of literature pertinent to low-precision back-propagation and training does not address the emergence of outliers in language models. As a result, our paper is dedicated to investigating the impact of outliers in low-precision integer training of language models. In the remainder of the paper, we demonstrate the significance of outlier features in the forward pass. Additionally, we establish that the forward pass can be entirely computed using integer arithmetic. Finally, we emphasize that handling outlier separately is only important in the forward pass. As for the back-propagation, all the parameters can remain in \texttt{INT8} format.

\section{\uppercase{Methodology}}

This section delves into our proposed methodology for mitigating outlier values in the low-precision training of language models. We consider keeping all the parameters in the back-propagation in \texttt{INT8} format while treating the outlier activation separately in the forward pass. We found that the outlier activation does not need to be treated differently (i.e. representing outliers in higher precision) in the back-propagation.

\subsection{Number Representation}

We employ the dynamic fixed-point format, also referred to as block floating-point \cite{williamson1991dynamically}, to convert floating-point numbers into integer data types. In this format, floating-point numbers are mapped into blocks of integer values, each assigned a unique scale. 

To perform this conversion, we utilize a linear fixed-point mapping function, which transforms a floating-point tensor $\mathbf{F}$ into a tensor of integers along with a single scale factor. The integer values are derived by rounding the floating-point mantissas, while the scale is determined as the maximum of the floating-point exponents within $\mathbf{F}$. The operational details of the linear fixed-point mapping are illustrated in the lower section of Figure \ref{fig:quant}.

To convert the fixed-point integers back into floating-point representation, we employ a non-linear inverse mapping function. This inverse mapping function converts integer values into normalized floating-point mantissas, associating each integer with its respective scale before packaging them into a floating-point number.

For a more comprehensive insight into the representation mapping functions, readers can refer to \cite{ghaffari2022integer}. It is worth noting that our approach deviates from existing methods by introducing various bit-widths for outlier activations in the fine-tuning of transformer-based language models. This strategy enables us to explore different bit-width configurations for handling outlier activations, ultimately facilitating the determination of the minimum memory band-width required for fine-tuning language models both in forward and backward propagations.

\subsection{Proposed methods}\label{sec:methodology}

In this subsection, we present two approaches designed to mitigate the impact of outliers in the context of low-precision language model fine-tuning. The first approach, the unified scale for outliers, seeks to provide a consistent scaling mechanism for outlier activations. The second approach, splitting outlier activations (Tiling), explores a novel strategy to isolate and manage outlier activations effectively, taking advantage of having two scaling factors for outliers.
Also note that in both approaches, we used a threshold $\gamma = 5$ to isolate the outlier activations for all the linear layers.

\subsubsection{Approach 1: Unified Scale for Outliers}
In this approach, we completely isolate the outlier activations and their corresponding weights and quantize them to \texttt{INT12} while the rest are quantized to \texttt{INT8}. Let us assume $\mathbf{X}$ denotes the activation tensor, and $Q()$ is the quantization function, then 

\begin{equation}
      Q(\mathbf{X}) = S_{\mathbf{x}} {\mathbf{X}} ^{\texttt{INT8}} + S_{\text{outlier}} {\mathbf{X}} ^{\texttt{INT12}}_{\text{outlier}}.
    \label{eq:method1}
\end{equation}

\subsubsection{Approach 2: Splitting Outlier Activations (Tiling)}

In the second approach, the value of outliers are split into two values $\hat{\mathbf{X}} ^{\texttt{INT8}}_{\text{outlier\_SP1}}$ and ${\mathbf{X}} ^{\texttt{INT8}}_{\text{outlier\_SP2}}$ and the quantization scheme is as follows

\begin{align}
      & Q(\mathbf{X}) = \\\nonumber
      & S_{\mathbf{x}} ( {\mathbf{X}} ^{\texttt{INT8}} +  {\mathbf{X}} ^{\texttt{INT8}} _{\text{outlier\_SP1}} )+ S_{\text{outlier}} {\mathbf{X}} ^{\texttt{INT8}}_{\text{outlier\_SP2}}.
    \label{eq:method2}
\end{align}

In this quantization scheme, we extract floating-point outlier activations $\mathbf{X}_{\text{outlier}}$  from original floating-point activations $\mathbf{X}$ using threshold $\gamma$ and then, split them as shown in the following equations,

\begin{align}
      & \mathbf{X}_{\text{outlier\_SP2}} = \lfloor \frac{\mathbf{X}_{\text{outlier}} + \gamma}{2\gamma} \rfloor \times 2\gamma \\\nonumber
      & \mathbf{X}_{\text{outlier\_SP1}} = \mathbf{X}_{\text{outlier}} - \mathbf{X}_{\text{outlier\_SP2}}.
    % \label{eq:sp1}
\end{align}
and then we quantize them to get ${\mathbf{X}} ^{\texttt{INT8}}_{\text{outlier\_SP1}}$ and ${\mathbf{X}} ^{\texttt{INT8}}_{\text{outlier\_SP2}}$. The benefit of this method is that we can keep the computation of forward pass completely in \texttt{INT8} format while treating the outlier separately from the no-outlier activation values.

\section{\uppercase{Theoretical Analysis}}

In this section, we delve into the implications of low-precision number formats on information preservation. The utilization of reduced bit-width representations in deep learning, while advantageous for efficiency and resource conservation, inevitably introduces the issue of information loss. We explore the nuances of this phenomenon and employ sensitivity analysis to quantify the extent to which information is altered or discarded in the transition from high precision to low precision.

Furthermore, we extend our investigation to consider distribution distances, such as the $\chi^2$-divergence and the Hammersley–Chapman–Robbins bound. 

\subsection{Information Loss in Low-precision Number Formats}

The concept of sample informativeness is a well-established notion within the field of statistics. For example, \cite{tukey1965part} introduced a dimensionless metric for measuring informativeness, which proves particularly valuable for our analysis. To measure the informativeness, \cite{tukey1965part} defines the concept of \textit{leverage} and \textit{linear sensitivity} as

\begin{align}
      \nonumber & \mathrm{lev}_{\theta}(X) = \frac{\partial}{\partial \theta} \mathbb{E}_{\theta}(X) \\
     & \mathrm{sens}_{\theta}(X) = \frac{(\mathrm{lev}_{\theta} (X))^2}{\mathbb{V}(X)},
    \label{eq:information}
\end{align}
where $\theta$ is the parameters of $X$ distribution.

Thus, we can re-write the equation \eqref{eq:information} in the low-precision number formats $\hat X$ if we consider a low precision number has a rounding error of $\delta$ in a way that $\hat X=  X + \delta$. Note that we assume $\delta$ and $X$ are independent random variables and $\mathbb{E}(\delta) = \varepsilon \simeq 0 $.

\begin{align}
      \nonumber & \frac{\mathrm{lev}_{\theta}(\hat X)}  {\mathrm{lev}_{\theta}(X)} \simeq 1 ~~~~~~ \text{s.t}~~~~ \mathbb{E}(\delta)\simeq 0 \\
      & \frac{\mathrm{sens}_{\theta}(\hat X)}{\mathrm{sens}_{\theta}(X)} = \frac{\mathbb{V}(X)}{\mathbb{V}(\hat X)} = \frac{\mathbb{V}(X)}{\mathbb{V}(X)+ \mathbb{V}(\delta)} \leq 1.
    \label{eq:info-loss}
\end{align}

Inequality \eqref{eq:info-loss} shows that in the case of unbiased rounding, low-precision representation always increases the variance and therefore decreases the informativeness of the sample.

It is noteworthy to mention that the sensitivity measure defined in equation \eqref{eq:information} is closely related to Hammersley-Chapman-Robbins lower-bound \cite{chapman1951minimum},

\begin{align}
      \nonumber & \frac{ \left ( \ \mathbb{E}(X) - \mathbb{E}(\hat X) \right )^2}{\mathbb{V}(\hat X)} \leq \\
      & \frac{\left (\mathbb{E}(X) - \mathbb{E}(\hat X)\right )^2}{\mathbb{V}(X)} \leq \chi^2(f_{X} || f_{\hat X}),
    \label{eq:hcr}
\end{align}
which provides a lower-bound for $\chi^2$-divergence of $X$ and $\hat X$ distributions. Note that $\chi^2$-divergence is a measure to quantify the divergence between two distributions and for distributions $P$ and $Q$ is defined as,

\begin{align}
    \chi^2(P|| Q)= \int (\frac{dP}{dQ} - 1)^2dQ.
    \label{eq:chi}
\end{align}

\begin{figure}[b]
\centering
\includegraphics[width=0.8\columnwidth]{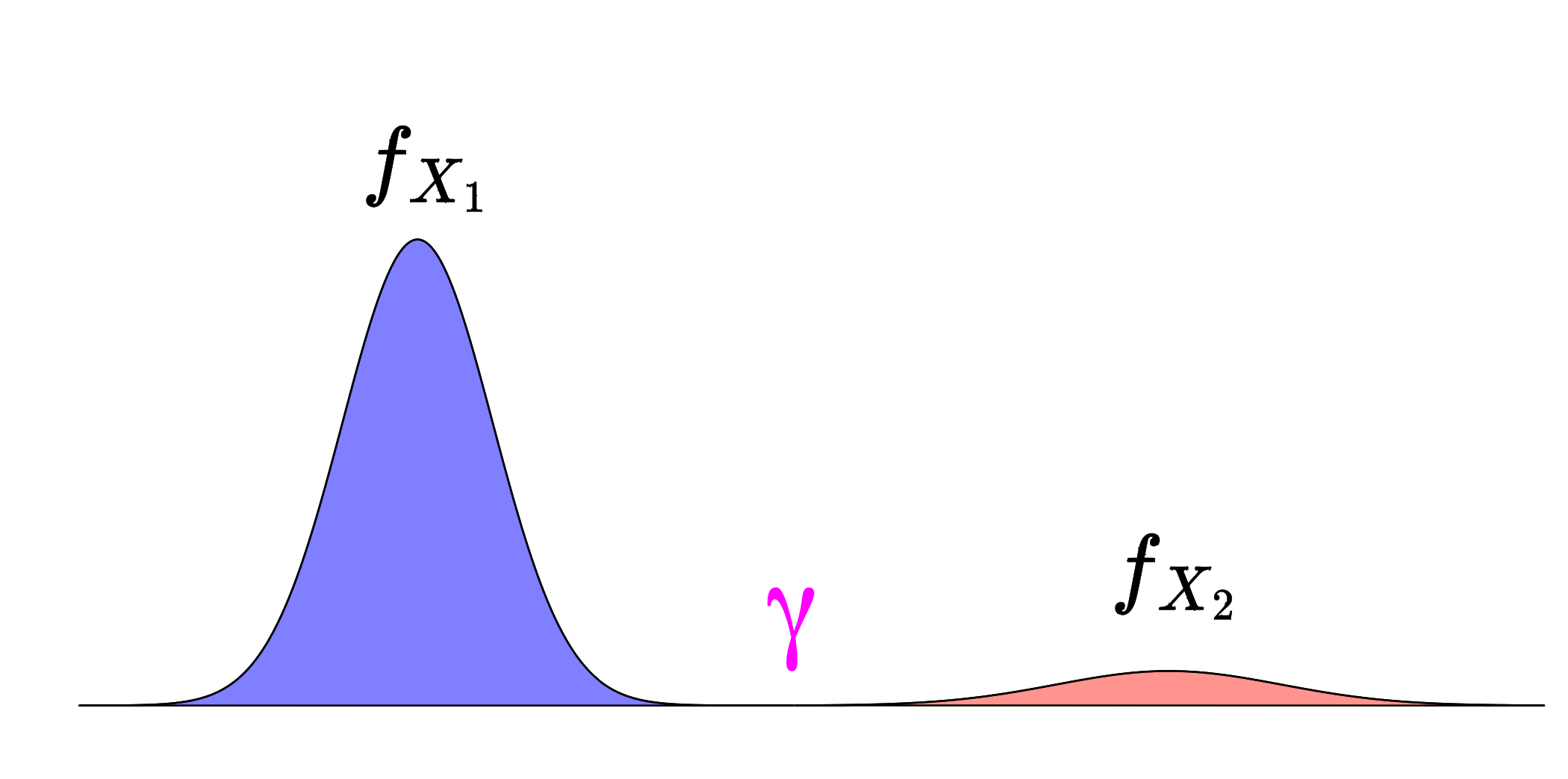}
\centering\caption{Outliers modeled as a mixture distribution.}
\label{fig:mixture}
\end{figure}

\subsection{Analysing Outlier Activations as a Mixture Distribution}\label{sec:mixture-informativeness}
Treating outlier activations separately as explained in Section \ref{sec:methodology} closely resembles having a mixture distribution as shown in Figure \ref{fig:mixture}. This means that we consider the outlier activations are samples that are drawn from a different distribution function than non-outlier activations. Let us assume the original distribution of $X$ as $f_{X}$ and a threshold $\gamma$ that separates outliers $f_{X_2}$ from the rest of activations $f_{X_1}$. define $F_{X}$ as the coefficients of $X$ such that

\begin{align}
    \nonumber & f_{X}(x) = pf_{X_1} (x) + (1-p)f_{X_2}(x),\\
    \nonumber & p = F_X(\gamma),\\
    \nonumber & f_{X_1} (x) = \frac{f_X(x) \I_{\{X\leq \gamma\}}}{F_X(\gamma)},\\
    & f_{X_2} (x) = \frac{f_X(x) \I_{\{X > \gamma\}}}{1-F_X(\gamma)}.
    \label{eq:mix1}
\end{align}

Let us further assume $Y=\I_{\{X\leq \gamma\}}$, then

\begin{align}
    \nonumber & \mathbb{E}(X) = \mathbb{E}\left ( \mathbb{E}(X|Y) \right ) \\
    & = p\mathbb{E}_{f_{X_1}} (X) + (1-p)\mathbb{E}_{f_{X_2}}(X),
    \label{eq:E_mix}
\end{align}

and,

\begin{align}
     \mathbb{E}(X^2) =  p\mathbb{E}_{f_{X_1}} (X^2) + (1-p)\mathbb{E}_{f_{X_2}}(X^2).
    \label{eq:E2_mix}
\end{align}

Furthermore, by subtracting the $p\mathbb{E}^2_{f_{X_1}} (X) + (1-p)\mathbb{E}^2_{f_{X_2}}(X)$ from \eqref{eq:E2_mix}
 and using Jensen's inequality we have

 \begin{align}
     \mathbb{V}(X) \geq  p\mathbb{V}_{f_{X_1}} (X) + (1-p)\mathbb{V}_{f_{X_2}}(X).
    \label{eq:V_mix}
\end{align}

\noindent\textbf{Remark 1.} The inequality \eqref{eq:V_mix} shows that weighted average of variances of distibutions is less than total variance of a mixture distribution. Therefore, treating outlier separately  reduces the variance and hence it increases the informativeness i.e. sensitivity according to equation \eqref{eq:information}.

\begin{table*}[!t]
\caption{Metric performance of integer fine-tuning of BERT on selected GLUE tasks. The reported metric for MRPC is accuracy and F1 score, for QNLI, MNLI, RTE, and SST-2 is accuracy, for STSB is the Pearson-Spearman correlation, and for CoLA is the Matthews correlation.}
\centering
\begin{tabular}{c|c|c|c|c|c|c|c}
                                                              & STSB  & QNLI  & MNLI  & SST-2 & RTE  & MRPC & CoLA \\\hline
% \textbf{Reported Parameters} & Pearson-Spearman correction  & Accuracy & Accuracy & Accuracy & Accuracy & Accuracy/F1 score & Matthews correlation \\ \hline
\small{\textbf{FP32}} & 87.6 & 89.9  & 83.5 & 91.9 & 61.7 & 78.7/85.3  & 55.3\\ 

\small{\textbf{FP16}} & 88.6 & 90.1  & 83.2 & 91.7 & 59.6 & 77.7/85.1  & 56.0
\\ \hdashline 

% \textbf{w:int8, out:fp32} & \textbf{85.5} & \textbf{89.9}  & 82.4 & 91.2 & 54.9 & 74.3/82.9 & \textbf{54.6} \\
\small{\textbf{Proposed Approach 1}} & 85.2  & 89.9  & 82.6 & 91.5 & 55.6  & 75.2/83.7 & 53.4 \\
\small{\textbf{Proposed Approach 2}} & 81.6 & 89.6 & 82.6 & 91.5& 59.2 & 74.3/83.5 & 52.2\\
\small{\textbf{\texttt{INT8} Untreated Outliers}} & 80.9  & 86.4  & 80.9 & 91.8 & 58.5 & 69.9/81.9 & 43.5 \\ 
% \textbf{w:int8, out:int16} & \textbf{85.5} & \textbf{89.9} & \textbf{82.9} & 91.4 & 58.5  & 74.5/83.1 & 54.5 \\

% \textbf{clamp(thr) w:int8, out: int8} & 83.6  & 86.6  & 82.1 & 89.8 & 54.2  & 74.8/82.8 & 36.1 \\
% \textbf{clamp(2*thr) w:int8, out: 2*int8} & 82.7  & 87.4  & 82.2 & 91.5 & 54.9  & 74.3/82.7 & 47.9 \\
\end{tabular}
\label{tab:glue}
\end{table*}

\begin{table}[!t]
\caption{Metric performance of fine-tuning BERT on SQuAD v1.1 and v2.0 datasets. For both datasets, the exact match metrics and F1 scores are reported.}
\centering
\begin{tabular}{c|c|c}
                    & SQuAD v1.1 & SQuAD v2 \\ \hline
% \textbf{Reported Parameters} & Exact match / F1 score & Exact match / F1 score \\ \hline 
\small{\textbf{FP32}} & 79.6/87.5 & 71.5/74.8 \\
\small{\textbf{FP16}} & 79.6/87.5 & 69.1/72.2 \\ \hdashline
\small{\textbf{Proposed Approach 1}}  & 76.2/85.2 & 67.7/71.2 \\
% \textbf{w:int8, out:fp32} & 76.2/85.0 & 68.1/71.5 \\
\small{\textbf{Proposed Approach 2}} & 74.9/84.1 & 65.5/69.0 \\
\small{\textbf{\texttt{INT8} Untreated Outliers}} & 69.8/80.2 & 60.9/64.6 \\
% \textbf{w:int8, out:int16} & 75.7/84.7 & \textbf{68.6/72.1} \\

% \textbf{clamp(thr) w:int8, out: int8} & 68.9/79.5 & 63.7/67.5 \\
% \textbf{clamp(2*thr) w:int8, out: 2*int8}  & 70.6/80.8 & 65.9/69.4 \\
\end{tabular}
% \caption{Metric performance of fine-tuning BERT on SQuAD v1.1 and v2.0 datasets. For both datasets, the exact match metrics and F1 scores are reported.}
\label{tab:squad}
\end{table}

\subsection{Informativeness of  Mixture Distribution in Low-Precision Number Formats}

In this section, we try to re-establish the results of Section \ref{sec:mixture-informativeness} for low-precision number formats. To do so, we need to show equation \eqref{eq:mix1} holds in the low-precision number format.

Let us consider the following low-precision representations, $\hat X= X + \delta$, $\hat X_1= X_1 + \delta$ and $\hat X_2= X_2 + \delta$, where $\delta$ is the rounding error and is independent from $X$. The moment generating function $m_{\hat X}$ is

 \begin{align}
     \nonumber & m_{\hat X}(t) = \mathbb{E}(e^{t \hat X})= \mathbb{E}(e^{tX})\mathbb{E}(e^{t \delta}) \\
     & = m_{X}(t) m_{\delta}(t).
    \label{eq:moment_generating_x_hat}
\end{align}
Now,
 \begin{align}
     \nonumber m_{ X}(t) & = \int_{-\infty}^{\infty} e^{tX}f_{X}(x) \\
     \nonumber & = p\int_{-\infty}^{\infty} e^{tX_1}f_{X_1}(x) \\
     \nonumber &+ (1-p) \int_{-\infty}^{\infty} e^{tX_2}f_{X_2}(x)\\
     & = pm_{X_1}(t) + (1-p)m_{X_2}(t) .
    \label{eq:moment_generating_x}
\end{align}
 and thus, using equations \eqref{eq:moment_generating_x_hat} and \eqref{eq:moment_generating_x},
 \begin{align}
     \nonumber m_{\hat X}(t) & = m_{X}(t) m_{\delta }(t) \\
     \nonumber & = p m_{X_1}(t) m_{\delta}(t) + (1-p)m_{X_2}(t) m_{\delta}(t)\\
     & = pm_{\hat X_1}(t) + (1-p)m_{\hat X_2}(t) .
    \label{eq:moment_generating_x_hat_2}
\end{align}
which means,

 \begin{align}
     f_{\hat X}(x) = pf_{\hat X_1} (x) + (1-p)f_{\hat X_2}(x).
    \label{eq:mix_dist_q}
\end{align}

\noindent\textbf{Remark 2.} Establishing equation \eqref{eq:mix_dist_q} confirms that inequality \eqref{eq:V_mix} holds for the low-precision regime and thus, treating outlier activations separately in low-precision reduces the quantization variance (i.e. quantization noise) and increases the informativeness.

\noindent\textbf{Remark 3.} The equation \eqref{eq:mix_dist_q} holds if the moement generative functions exist. This is a valid assumption since the distribution of activation has bounded support.

\section{\uppercase{Experimental results}}
\subsection{Experiment Setup}
We conducted fine-tuning on the BERT base model across a series of downstream tasks to facilitate a performance comparison between our integer fine-tuning method and the \texttt{FP16} and \texttt{FP32} fine-tuning approaches. The fine-tuning process encompassed specific tasks selected from the GLUE benchmark \cite{wang2018glue}, in addition to the Stanford Question Answering Datasets, specifically SQuAD v1.1 and SQuAD v2.0 \cite{rajpurkar2016squad}.

Each fine-tuning setup was standardized with identical hyper-parameters and an equivalent number of training epochs. To ensure result stability, reported metrics represent the average of five runs, each initialized with a different random seed to mitigate the impact of random variations.

Our fine-tuning experiments were executed using the fine-tuning scripts provided by the Hugging Face library \cite{wolf2019huggingface}. In the case of GLUE experiments, fine-tuning spanned five epochs, with a learning rate set to $2\times 10^{-5}$, and a per-device fine-tuning batch size of 32. Meanwhile, the fine-tuning of BERT on the SQuAD datasets comprised two epochs, with a learning rate of $5 \times 10^{-5}$ and a per-device fine-tuning batch size of 12. Notably, all experiments were conducted on a computational infrastructure consisting of eight NVIDIA V100 GPUs, each equipped with 32 gigabytes of VRAM. 

\subsection{Results}
The results of fine-tuning BERT base on GLUE benchmark and SQuAD datasets are presented in Table \ref{tab:glue} and Table \ref{tab:squad} respectively. Our proposed solution significantly improves the robustness of low-precision fine-tuning for BERT on GLUE and SQuAD datasets. Comparing the results of the proposed approaches with \texttt{INT8} fine-tuning with untreated outliers shows that representing outliers separately almost always improves the fine-tuning performance of the model. Additionally, we emphasize again that, the results  presented in this paper comprise of having both back-propagation and forward-propagation in low-precision number formats and show that low-precision arithmetic are promising avenue to reduce the computational complexity of language models.

\section{\uppercase{Conclusion}}

This paper explored means to mitigate the outlier activations in low-precision language model fine-tuning. We have introduced a novel methodology for mitigating the challenges posed by outlier activations, offering effective approaches as effective approaches to enhance the stability of the fine-tuning phase in low precision number format where gradients, weights, and activations are in \texttt{INT8} format.
Additionally, we provided a theoretical analysis to understand the intricacies of information loss in low-precision number formats. Our sensitivity analysis has unveiled the trade-offs between variance and informativeness while considering distribution distances like the $\chi^2$-divergence and the Hammersley–Chapman–Robbins bound has deepened our insights into these transformations.
In a landscape where the deployment of large language models is increasingly resource-constrained, our work contributes to the ongoing efforts to make these models more accessible and efficient. By addressing the challenges of outliers and information loss, we pave the way for the continued evolution of low-precision back-propagation in language model fine-tuning. Our findings not only have implications for natural language processing but also hold relevance for broader applications across data analysis and machine learning.

\bibliographystyle{apalike}
{\small
\bibliography{example}}

\begin{thebibliography}{}

\bibitem[Chapman and Robbins, 1951]{chapman1951minimum}
Chapman, D.~G. and Robbins, H. (1951).
\newblock Minimum variance estimation without regularity assumptions.
\newblock {\em The Annals of Mathematical Statistics}, pages 581--586.

\bibitem[Dettmers et~al., 2022]{dettmers2022llm}
Dettmers, T., Lewis, M., Belkada, Y., and Zettlemoyer, L. (2022).
\newblock Llm. int8 (): 8-bit matrix multiplication for transformers at scale.
\newblock {\em arXiv preprint arXiv:2208.07339}.

\bibitem[Dettmers et~al., 2023]{dettmers2023spqr}
Dettmers, T., Svirschevski, R., Egiazarian, V., Kuznedelev, D., Frantar, E., Ashkboos, S., Borzunov, A., Hoefler, T., and Alistarh, D. (2023).
\newblock Spqr: A sparse-quantized representation for near-lossless llm weight compression.
\newblock {\em arXiv preprint arXiv:2306.03078}.

\bibitem[Dettmers and Zettlemoyer, 2023]{dettmers2023case}
Dettmers, T. and Zettlemoyer, L. (2023).
\newblock The case for 4-bit precision: k-bit inference scaling laws.
\newblock In {\em International Conference on Machine Learning}, pages 7750--7774. PMLR.

\bibitem[Frantar et~al., 2022]{frantar2022gptq}
Frantar, E., Ashkboos, S., Hoefler, T., and Alistarh, D. (2022).
\newblock Gptq: Accurate post-training quantization for generative pre-trained transformers.
\newblock {\em arXiv preprint arXiv:2210.17323}.

\bibitem[Ghaffari et~al., 2022]{ghaffari2022integer}
Ghaffari, A., Tahaei, M.~S., Tayaranian, M., Asgharian, M., and Partovi~Nia, V. (2022).
\newblock Is integer arithmetic enough for deep learning training?
\newblock {\em Advances in Neural Information Processing Systems}, 35:27402--27413.

\bibitem[Li et~al., 2023]{li2023norm}
Li, L., Li, Q., Zhang, B., and Chu, X. (2023).
\newblock Norm tweaking: High-performance low-bit quantization of large language models.
\newblock {\em arXiv preprint arXiv:2309.02784}.

\bibitem[Lin et~al., 2023]{lin2023awq}
Lin, J., Tang, J., Tang, H., Yang, S., Dang, X., and Han, S. (2023).
\newblock Awq: Activation-aware weight quantization for llm compression and acceleration.
\newblock {\em arXiv preprint arXiv:2306.00978}.

\bibitem[Micikevicius et~al., 2017]{micikevicius2017mixed}
Micikevicius, P., Narang, S., Alben, J., Diamos, G., Elsen, E., Garcia, D., Ginsburg, B., Houston, M., Kuchaiev, O., Venkatesh, G., et~al. (2017).
\newblock Mixed precision training.
\newblock {\em arXiv preprint arXiv:1710.03740}.

\bibitem[Rajpurkar et~al., 2016]{rajpurkar2016squad}
Rajpurkar, P., Zhang, J., Lopyrev, K., and Liang, P. (2016).
\newblock Squad: 100,000+ questions for machine comprehension of text.
\newblock {\em arXiv preprint arXiv:1606.05250}.

\bibitem[Tayaranian et~al., 2023]{hosseini2023towards}
Tayaranian, M., Ghaffari, A., Tahaei, M.~S., Rezagholizadeh, M., Asgharian, M., and Nia, V.~P. (2023).
\newblock Towards fine-tuning pre-trained language models with integer forward and backward propagation.
\newblock In {\em Findings of the Association for Computational Linguistics: EACL 2023}, pages 1867--1876.

\bibitem[Tukey, 1965]{tukey1965part}
Tukey, J.~W. (1965).
\newblock Which part of the sample contains the information?
\newblock {\em Proceedings of the National Academy of Sciences}, 53(1):127--134.

\bibitem[Wang et~al., 2018]{wang2018glue}
Wang, A., Singh, A., Michael, J., Hill, F., Levy, O., and Bowman, S.~R. (2018).
\newblock Glue: A multi-task benchmark and analysis platform for natural language understanding.
\newblock {\em arXiv preprint arXiv:1804.07461}.

\bibitem[Wei et~al., 2022]{wei2022outlier}
Wei, X., Zhang, Y., Zhang, X., Gong, R., Zhang, S., Zhang, Q., Yu, F., and Liu, X. (2022).
\newblock Outlier suppression: Pushing the limit of low-bit transformer language models.
\newblock {\em Advances in Neural Information Processing Systems}, 35:17402--17414.

\bibitem[Williamson, 1991]{williamson1991dynamically}
Williamson, D. (1991).
\newblock Dynamically scaled fixed point arithmetic.
\newblock In {\em [1991] IEEE Pacific Rim Conference on Communications, Computers and Signal Processing Conference Proceedings}, pages 315--318. IEEE.

\bibitem[Wolf et~al., 2019]{wolf2019huggingface}
Wolf, T., Debut, L., Sanh, V., Chaumond, J., Delangue, C., Moi, A., Cistac, P., Rault, T., Louf, R., Funtowicz, M., et~al. (2019).
\newblock Huggingface's transformers: State-of-the-art natural language processing.
\newblock {\em arXiv preprint arXiv:1910.03771}.

\bibitem[Xiao et~al., 2023]{xiao2023smoothquant}
Xiao, G., Lin, J., Seznec, M., Wu, H., Demouth, J., and Han, S. (2023).
\newblock Smoothquant: Accurate and efficient post-training quantization for large language models.
\newblock In {\em International Conference on Machine Learning}, pages 38087--38099. PMLR.

\bibitem[Yuan et~al., 2023]{yuan2023rptq}
Yuan, Z., Niu, L., Liu, J., Liu, W., Wang, X., Shang, Y., Sun, G., Wu, Q., Wu, J., and Wu, B. (2023).
\newblock Rptq: Reorder-based post-training quantization for large language models.
\newblock {\em arXiv preprint arXiv:2304.01089}.

\bibitem[Zhang et~al., 2020]{zhang2020fixed}
Zhang, X., Liu, S., Zhang, R., Liu, C., Huang, D., Zhou, S., Guo, J., Guo, Q., Du, Z., Zhi, T., et~al. (2020).
\newblock Fixed-point back-propagation training.
\newblock In {\em Proceedings of the IEEE/CVF Conference on Computer Vision and Pattern Recognition}, pages 2330--2338.

\bibitem[Zhao et~al., 2021]{zhao2021distribution}
Zhao, K., Huang, S., Pan, P., Li, Y., Zhang, Y., Gu, Z., and Xu, Y. (2021).
\newblock Distribution adaptive int8 quantization for training cnns.
\newblock In {\em Proceedings of the AAAI Conference on Artificial Intelligence}, volume~35, pages 3483--3491.

\bibitem[Zhao et~al., 2019]{zhao2019improving}
Zhao, R., Hu, Y., Dotzel, J., De~Sa, C., and Zhang, Z. (2019).
\newblock Improving neural network quantization without retraining using outlier channel splitting.
\newblock In {\em International conference on machine learning}, pages 7543--7552. PMLR.

\bibitem[Zhu et~al., 2020]{zhu2020towards}
Zhu, F., Gong, R., Yu, F., Liu, X., Wang, Y., Li, Z., Yang, X., and Yan, J. (2020).
\newblock Towards unified int8 training for convolutional neural network.
\newblock In {\em Proceedings of the IEEE/CVF Conference on Computer Vision and Pattern Recognition}, pages 1969--1979.

\end{thebibliography}

% \section*{\uppercase{Appendix}}

% If any, the appendix should appear directly after the
% references without numbering, and not on a new page. To do so please use the following command:
% \textit{$\backslash$section*\{APPENDIX\}}

\end{document}